\documentclass[conference]{IEEEtran}
\IEEEoverridecommandlockouts
\usepackage{cite}
\usepackage{amsmath,amssymb,amsfonts}
\usepackage{algorithm}
\usepackage{algorithmic}
\usepackage{graphicx}
\usepackage{textcomp}
\usepackage{xcolor}
\def\BibTeX{{\rm B\kern-.05em{\sc i\kern-.025em b}\kern-.08em
    T\kern-.1667em\lower.7ex\hbox{E}\kern-.125emX}}

\usepackage{eso-pic}

\AddToShipoutPictureFG*{%
  \AtPageUpperLeft{%
    \raisebox{-1.2cm}{%
      \makebox[\paperwidth]{%
        \color{black}\footnotesize
        This work has been submitted to the IEEE for possible publication.
        Copyright may be transferred without notice, after which this version
        may no longer be accessible.
      }
    }
  }
}

\begin{document}

\title{Crane Lowering Guidance Using a Attachable Camera Module for Driver Vision Support\\
\thanks{This work is supported by the Korea Agency for Infrastructure Technology Advancement(KAIA) grant funded by the Ministry of Land, Infrastructure and Transport (Grant No. 2610000426)}
}

\author{\IEEEauthorblockN{1\textsuperscript{st} HyoJae Kang$^\dagger$}
\IEEEauthorblockA{\textit{Department of Interdisciplinary} \\
\textit{Robot Engineering Systems} \\
\textit{Hanyang University}\\
Ansan, South Korea \\
majaae5@hanyang.ac.kr\\
$^\dagger$ These authors are contributed \\equally to this work.}

\and
\IEEEauthorblockN{2\textsuperscript{nd} SunWoo Ahn$^\dagger$}
\IEEEauthorblockA{\textit{Department of}\\
\textit{Smart Construction Engineering}\\
\textit{Hanyang University}\\
Ansan, South Korea \\
sunoo0210@hanyang.ac.kr}

\and
\IEEEauthorblockN{3\textsuperscript{rd} InGyu Choi}
\IEEEauthorblockA{\textit{Department of Robot Engineering}\\
\textit{Hanyang University}\\
Ansan, South Korea \\
chkchk321@hanyang.ac.kr}

\and
\IEEEauthorblockN{4\textsuperscript{th} GeonYeong Go}
\IEEEauthorblockA{\textit{Department of}\\
\textit{Smart Construction Engineering}\\
\textit{Hanyang University}\\
Ansan, South Korea \\
vontte21@hanyang.ac.kr}

\and
\IEEEauthorblockN{5\textsuperscript{th} KunWoo Son}
\IEEEauthorblockA{\textit{School of Smart}\\
\textit{Convergence Engineering} \\
\textit{Hanyang University}\\
Ansan, South Korea \\
skxwoo@hanyang.ac.kr}

\and
\IEEEauthorblockN{6\textsuperscript{th} Min-Sung Kang*}
\IEEEauthorblockA{\textit{School of Smart}\\
\textit{Convergence Engineering} \\
\textit{Hanyang University}\\
Ansan, South Korea \\
wowmecha@hanyang.ac.kr}
*Corresponding author
~\\
}

\maketitle

\begin{abstract}
Cranes have long been essential equipment for lifting and placing heavy loads in construction projects. This study focuses on the lowering phase of crane operation, the stage in which the load is moved to the desired location. During this phase, a constant challenge exists: the load obstructs the operator's view of the landing point. As a result, operators traditionally have to rely on verbal or gestural instructions from ground personnel, which significantly impacts site safety. To alleviate this constraint, the proposed system incorporates a attachable camera module designed to be attached directly to the load via a suction cup. This module houses a single-board computer, battery, and compact camera. After installation, it streams and processes images of the ground directly below the load in real time to generate installation guidance. Simultaneously, this guidance is transmitted to and monitored by a host computer. Preliminary experiments were conducted by attaching this module to a test object, confirming the feasibility of real-time image acquisition and transmission. This approach has the potential to significantly improve safety on construction sites by providing crane operators with an instant visual reference of hidden landing zones.
\end{abstract}


\section{Introduction}

Lifting heavy objects is essential on construction sites, and this work is carried out using a variety of cranes, including mobile cranes, tower cranes, and gantry cranes. The operations of a crane operated directly by a human can be broadly divided into three parts: (1) The trolley is moved above the object's starting point, and the worker secures the object to a sling or hook. (2) After lifting the object, the crane operates to move it above the desired target point. (3) The object is gradually lowered in coordination with a worker standing above the target point. Looking at this series of crane operations, we can see that smooth communication with on-site workers is essential, as the load can sometimes obscure the target point during the lowering process.

Machine guidance and machine control technologies have been researched for the many construction machines required on construction sites, with the aim of supporting manual work, unmanning, and automating them. Machine guidance refers to technology that assists workers in operating machines, while machine control refers to technology that automatically controls machines without worker intervention\cite{c1,c2}. These technologies have also been introduced to cranes, and can be mainly categorized into crane safety alarm systems and crane control technologies.

Research on crane safety includes safety monitoring using cameras installed on the crane boom\cite{c3}, dynamic collision pre-warning based on trajectories\cite{c4}, worker detection around tower cranes and collision risk assessment\cite{c5}, collision probability assessment between tower crane work and workers\cite{c6}, real-time collision risk assessment by detecting workers and hooks and applying safety rules based on distance thresholds\cite{c7}, potential collision prediction using drone footage\cite{c8}, vision systems for tower crane monitoring and risk warnings combining distance thresholds and trajectory analysis\cite{c9}, and potential collision prediction analysis based on digital twins\cite{c10}.

Research into crane automation includes anti-sway research to suppress swaying when a crane moves an object\cite{c11,c12}, research into skew control related to bridge torsion\cite{c13,c14}, and research into cooperative control of multiple cranes\cite{c15}.

Despite the application of various research methods, accidents still occur frequently at worksites. Especially in crane lowering, blind spots are created by the suspended object. Importantly, the presence of workers below the object makes safety extremely vulnerable. Many studies have used vision systems to monitor the crane's working environment, but despite this, sufficient improvements to these blind spots have not yet been achieved. In particular, when large-scale components need to be lifted, such as at a precast concrete construction site, not only the installation location of the module is important, but a complex situation arises in which workers are also located directly below the object being lifted.

Therefore, the purpose of this study is to monitor the vertically downward area of an object being lifted by a crane and provide guidance on where the object will be placed at the target location. To achieve this, a module equipped with a camera and a laser pointer was created, and after processing the data using an single board computer(SBC) inside the module, the video data was sent to a host PC. The module is easily attachable using a suction cup, and can provide guidance by attaching it to the side of the object being lifted. Furthermore, since data can be sent from each side depending on the number of modules, guidance can be provided from multiple directions. This makes it possible to obtain data from below the object being lifted, improving safety on site. In particular, the lifting objects targeted in this study were limited to rectangular shapes with edges.

The next section describes the design of the camera module, and Section 3 presents the image processing data. Section 4 presents experimental results of providing guidance functions by attaching the camera to the side of an object suspended in the air. Section 5 presents conclusions and limitations, and Section 6 discusses future research topics.

\section{Module Design}

The camera module consists of a camera (Arducam B0191) for capturing images, a Laser Pointer (WAT-L335, wavelength 532 nm, 50 mW) a single-board computer (Raspberry Pi 4B) for image processing and data transmission, a battery for power supply, a connection part for the lifting object, and a frame. Fig. 1 shows the hardware configuration of the camera module.

First, the connection part is designed so that when attached to the side of the lifting object, the camera is oriented so that it can capture images vertically downward. The connection part also uses a suction cup, which allows the module to be easily attached by pressing it against the member. While this study focused on rectangular box shapes and used suction cups, magnets can also be used for container boxes or other magnetic objects, and the attachment method can be changed. The frame is made of PLA material using a 3D printer, and is lightweight, making it easy to install.

\begin{figure}[!t]
    \centering
    \includegraphics[width=0.99\columnwidth]{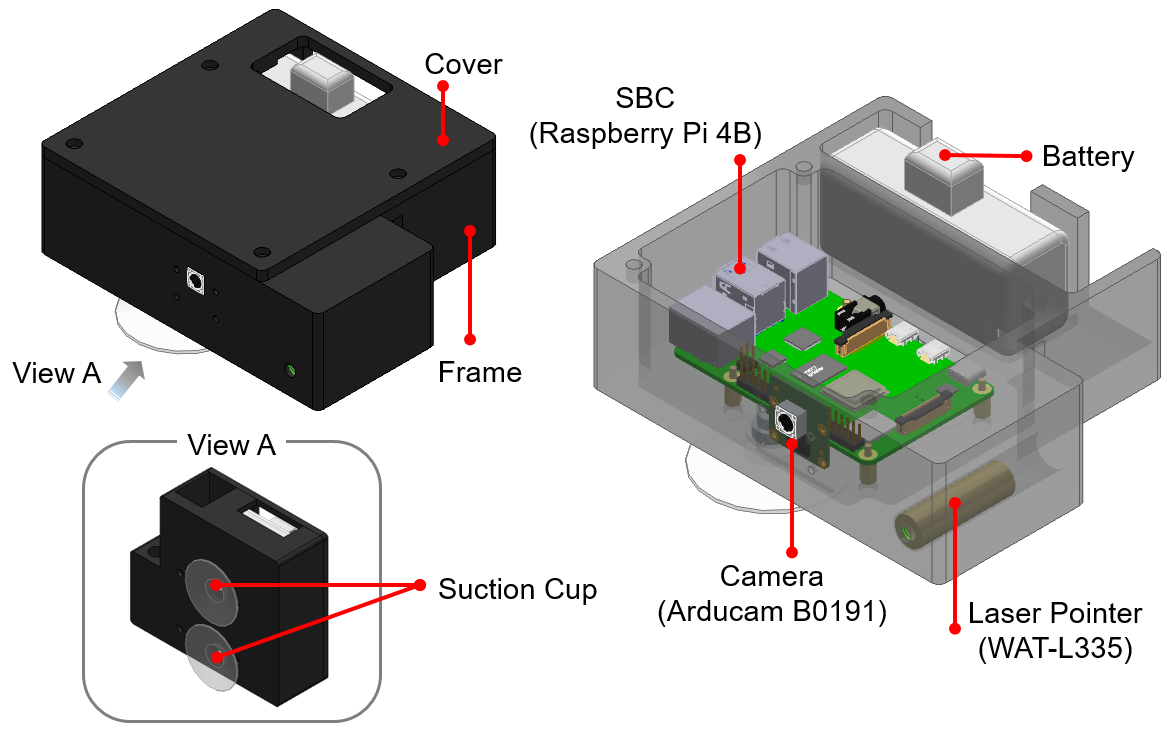}
    \caption{Detailed Design of the Camera Module}
    \label{fig:placeholder}
\end{figure}

Next, Fig. 2 shows the design dimensions of the camera module and the dimensions when the module is attached to an object surface. The module measures 50 mm × 150 mm × 127 mm. When the suction cup is pressed to the maximum extent possible to attach the module to the object surface (black, dotted line), the height of the end of the suction cup (blue, dashed line) from the object surface is 7.8 mm, the distance from the object surface to the center of the laser pointer (green, dashed line) is 19.8 mm, and the distance from the object surface to the center of the camera lens (navy blue, dashed line) is 36.8 mm.

\begin{figure}[!t]
    \centering
    \includegraphics[width=0.99\columnwidth]{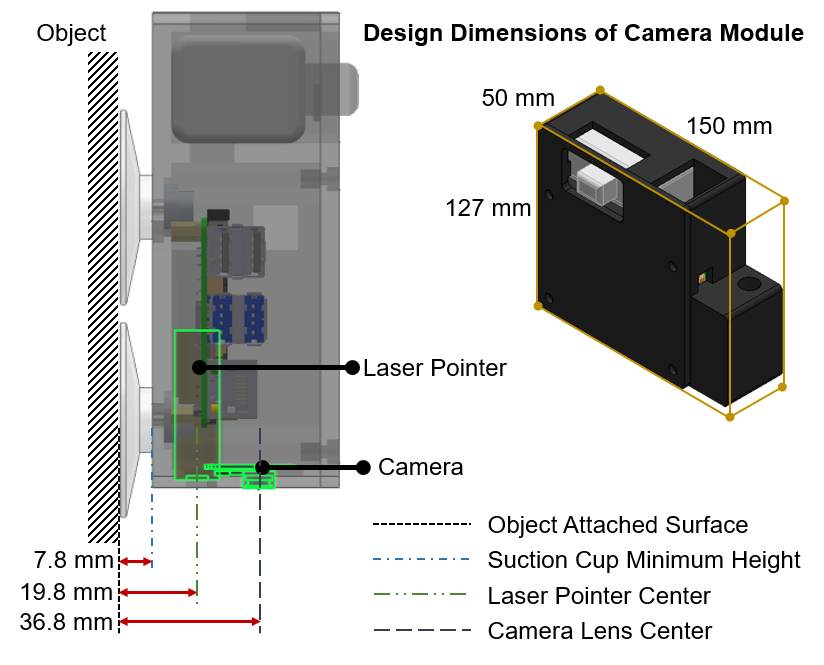}
    \caption{Position of the camera and laser pointer when the module is attached to the object and design dimensions of the module}
    \label{fig:placeholder}
\end{figure}

Fig. 3 shows the communication flow and power line of the electrical components used in the camera module. First, data acquired by the camera is sent to the SBC via serial communication and processed by an algorithm built into the Raspberry Pi 4B. The data is then transferred to the host PC via wifi communication. This allows the host PC to receive the data and perform monitoring. Laser pointers are connected only to a power line and are designed to turn on immediately when electricity is applied.

\begin{figure}[!t]
    \centering
    \includegraphics[width=0.99\columnwidth]{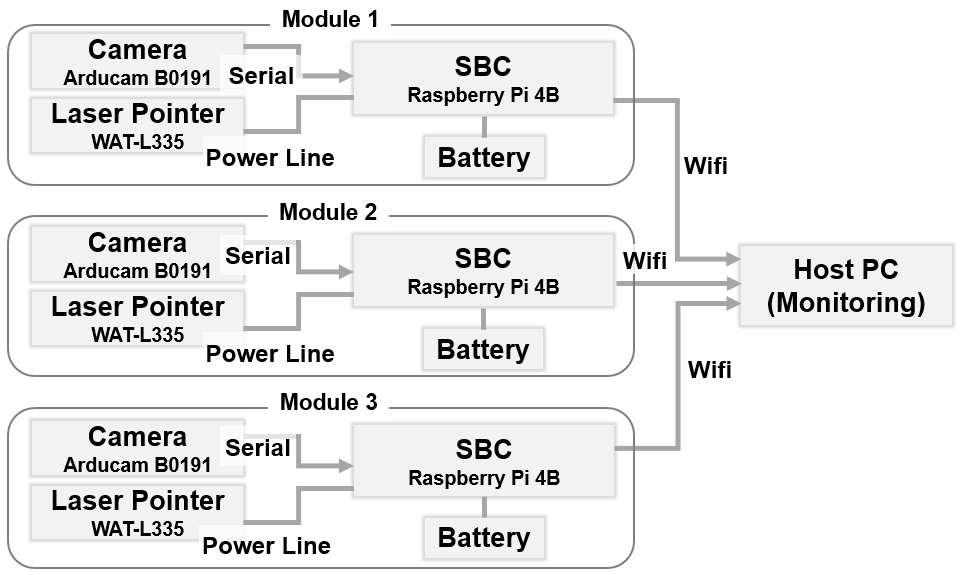}
    \caption{Communication Flow and Power Line of the Camera Modules}
    \label{fig:placeholder}
\end{figure}

In this study, a total of three modules were fabricated, each configured to transmit processed video to a host PC. Therefore, it is possible to add modules up to the number that the host PC can receive, and in this study, modules were installed in up to three corners.

\section{Guidance System}

This section details the location of the camera module on the component, the data processing method, and the guidance provided, as presented in Section 2. In particular, the location of the camera module on the component is an important factor, as it is related to the data processing standards.

First, the proposed camera module is based on a cuboid component. Examples of lifting targets include loaded pallets, precast concrete modules, steel boxes, steel beams, and containers. Fig. 4 shows the positioning of the camera module on a component with the proposed shape. The key point in attachment is to ensure that the edges of the component are captured within the camera's field of view. Because edge information is required for data processing. Data processing is difficult if the module is installed in a position where edges are not captured, or if the component has a shape without edges, such as a cylinder. Fig. 4 also shows a blind area, which is used to compensate for occluded areas and provide guidance for the placement of the object on the ground in the image. Finally, since the camera image alone provides insufficient information about the ground, making it difficult to predict the placement position, a laser pointer is used to generate markers on the ground.

\begin{figure}[!t]
    \centering
    \includegraphics[width=0.85\columnwidth]{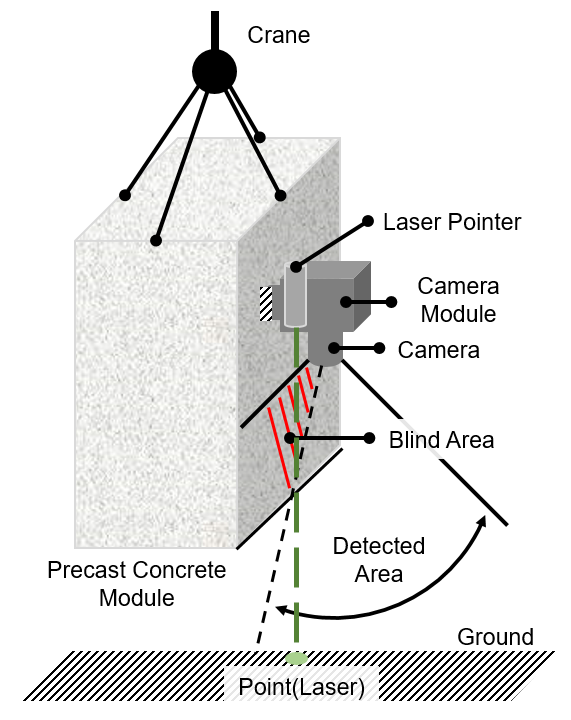}
    \caption{Guidance concept for the landing position of the lifted object}
    \label{fig:placeholder}
\end{figure}

Fig. 5 shows an example of an image captured when the camera module is attached as shown in Fig. 4, where only one edge is captured. The left image in Fig. 5 shows the position of the laser pointer on the ground and the captured ground area when the camera module is attached, as in Fig. 4. On the other hand, the right image in Fig. 5 is an example of a captured image, where it can be seen that the edge is distorted.

\begin{figure}[!t]
    \centering
    \includegraphics[width=0.95\columnwidth]{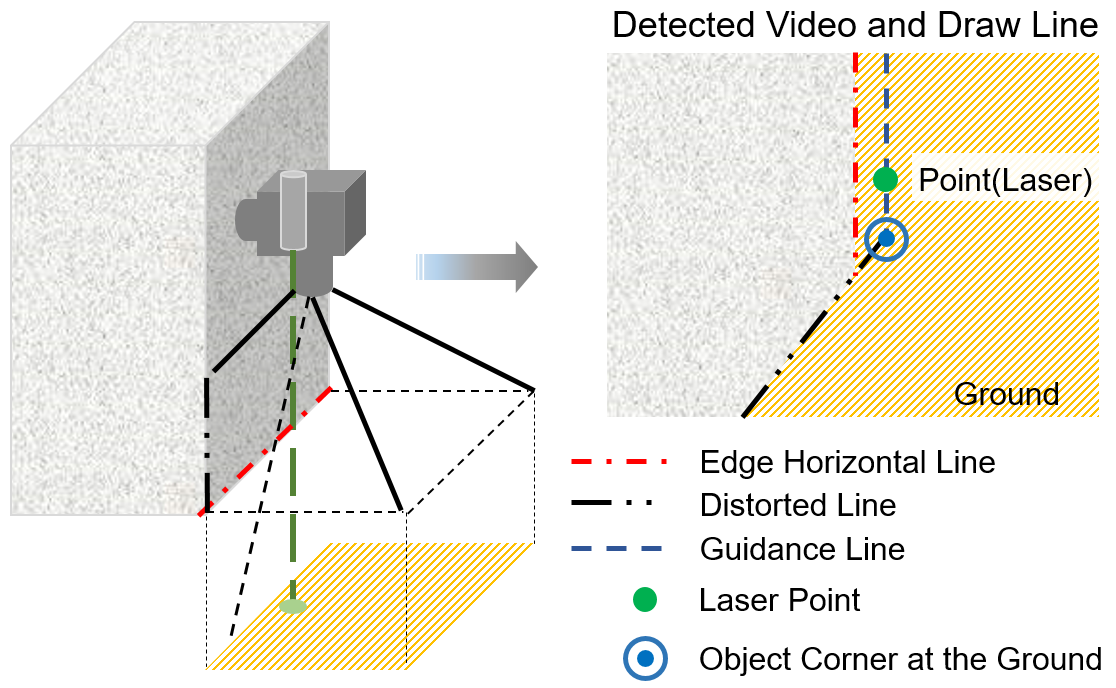}
    \caption{Concept of guidance for placement of lifted objects}
    \label{fig:placeholder}
\end{figure}

This explains how to use this to draw guidance for the position where a lifted object will be placed on the ground. First, draw a distorted line(black, dash-double dotted line), and extend this line from the edge. Next, draw a line that is parallel to the undistorted horizontal line of the lifted object (red, dash-single dotted line) and passes through the point indicated by the laser pointer (green, point). The point where these two lines intersect is the position where the corner of the object actually touches the ground (blue, double circle), and can be shown as a guidance Line (blue, dashed line).

Based on this research, the line detection and selection algorithm for an input image is shown in Fig. 6 - Algorithm 1. Algorithm 1 uses grayscale conversion, Hough transform, Canny edge detection, and Gaussian blurring from the OpenCV library. The parameters used in this paper are $\lvert \theta_{horiz\_max} \rvert \leq 10^\circ$ and $20^\circ < \lvert \theta_{diag\_range} \rvert < 70^\circ$ was applied. The proposed algorithm consists of detecting representative horizontal and diagonal lines from the input image and extending them to the image boundary to finally output two lines. First, the input RGB image is converted to grayscale. Then, Gaussian blurring is applied to reduce noise, and Canny edge detection is performed. A Hough transform is performed based on the resulting edge image $E$ to detect a set of line segment candidates $S$. For each detected line segment, the length, angle, and position features are calculated to generate a set of horizontal line candidates $H$ and a set of diagonal line candidates $D$. Each candidate line is evaluated using SCORE, and a representative line is selected. After that, the selected line is extended to the image boundary after validity verification.

\begin{figure}
    \centering
    \includegraphics[width=0.9\columnwidth]{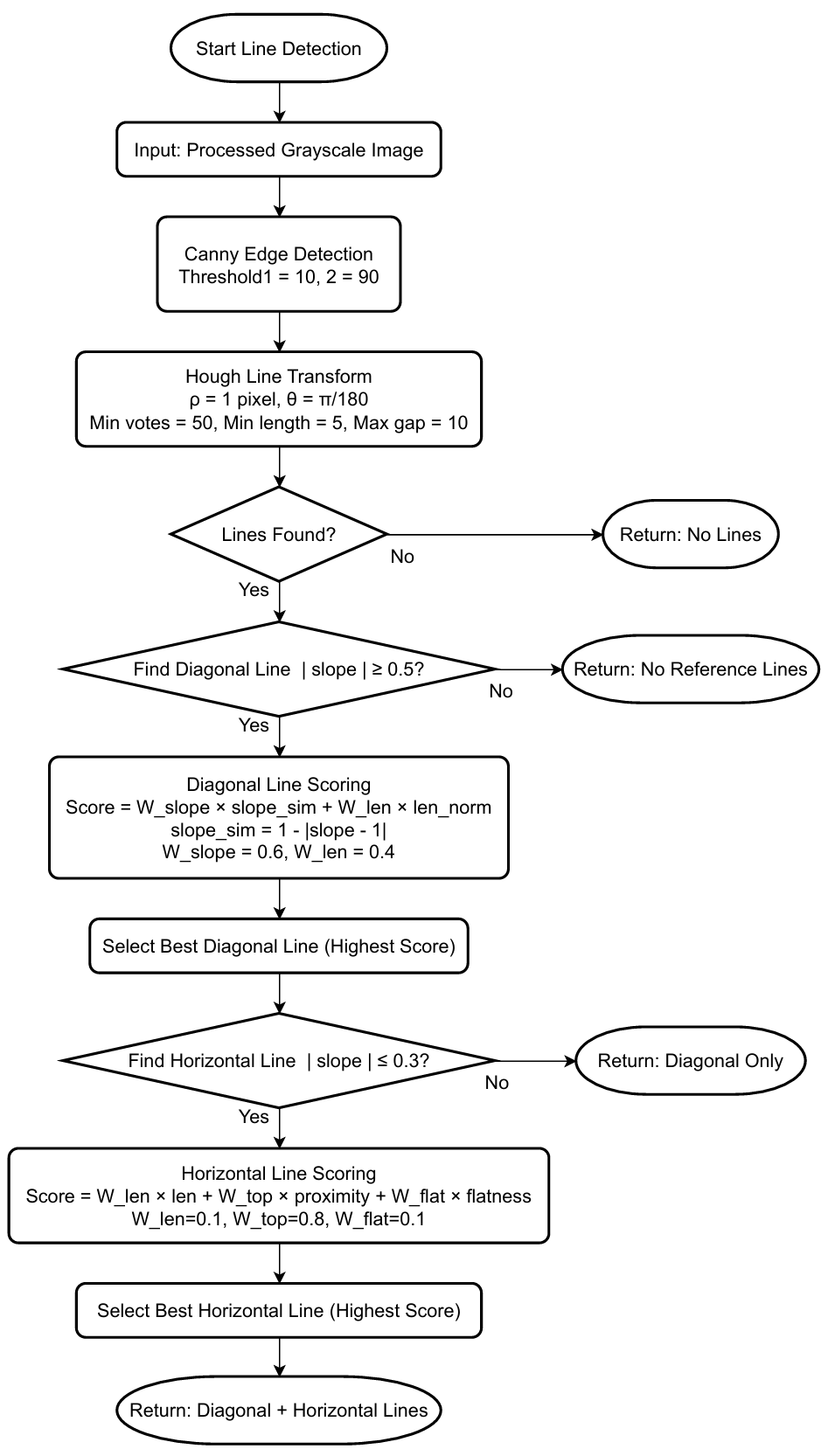}
    \caption{Algorithm 1: Line detecting and selecting}
    \label{fig:placeholder}
\end{figure}

Algorithm 2 as shown in Fig. 7 aims to detect the center coordinates of a green laser spot in an image. The OpenCV libraries used are Gaussian Blur and Contour Detection. Laser pointers generally have small, circular shapes and high saturation and brightness, making their detection robust even with simple methods that exploit their color and brightness characteristics. In this study, we extracted green laser pointer points by combining mask generation based on the HSV color space with morphological filtering. Specifically, the input video is converted to the HSV color space and binary masks are generated for two ranges corresponding to greens (the core green region and the bright green region). These two masks are then combined to identify potential laser candidate regions, after which morphological operations and Gaussian blurring are applied to remove noise. Contour detection is then performed, and only candidates that meet certain area criteria are retained. Finally, the candidate with the largest area is selected as the representative laser point, and its center coordinates are calculated using moment-based calculations. If no candidates exist, no results are returned.

\begin{figure}
    \centering
    \includegraphics[width=0.8\columnwidth]{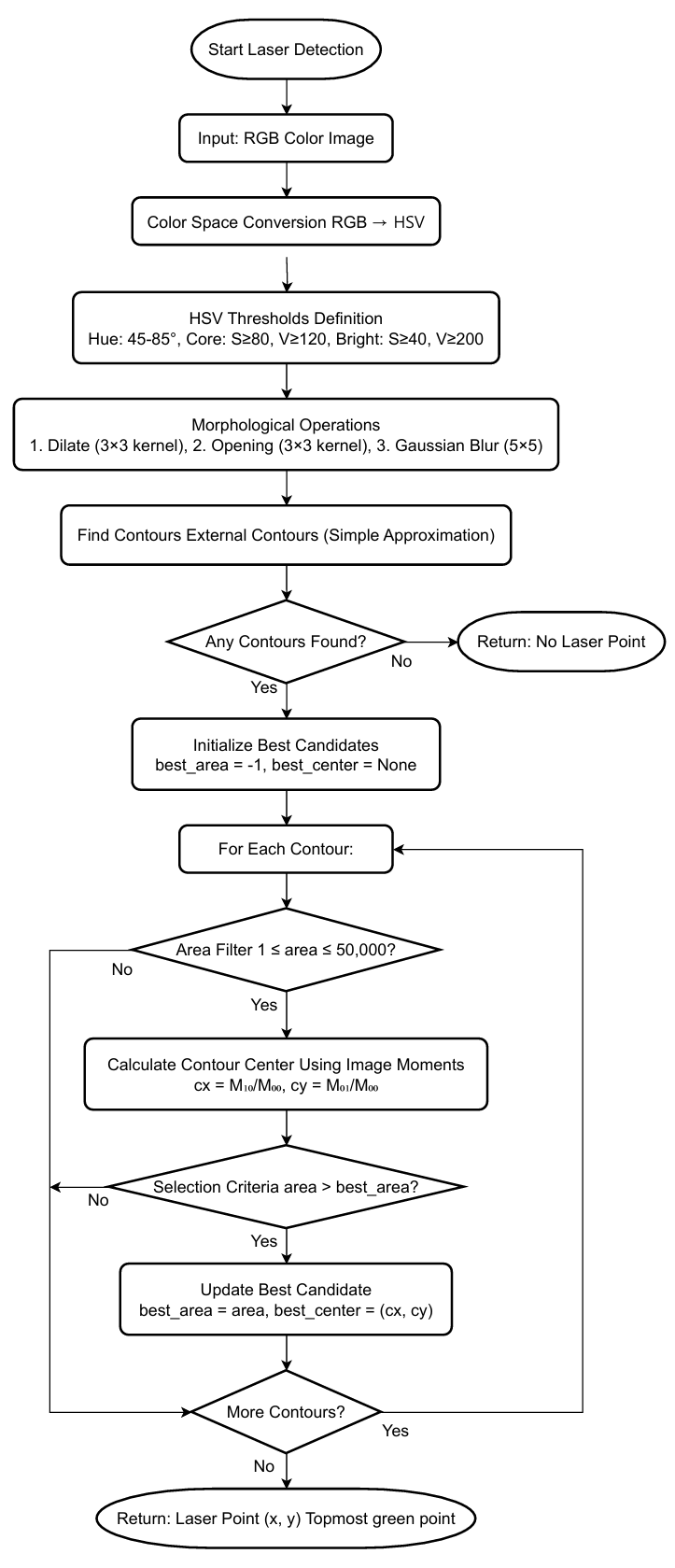}
    \caption{Algorithm 2: Laser spot detecting}
    \label{fig:placeholder}
\end{figure}

Based on the two presented algorithms, a camera module is used to provide the guidance line of the object along with real-time video.

\section{Experimental Validation}

The experiment was conducted in an indoor environment with the aim of verifying whether the camera module could perform the proposed functions. The experiment was conducted by attaching the camera module to a frame as the object. The frame was not actually lifted, but a wall was used to confirm the guidance function. It should be noted that the experiment was conducted with horizontal manner instead of using suspended frame, which does not fully replicate a real crane operation. Therefore, the results should be interpreted as a preliminary validation ratheer than a direct field application.

In the experiment, three modules were attached to process information about three corners. Furthermore, since it was difficult to attach the module directly to the frame using suction cups, an acrylic plate was added to the mounting surface and the module was attached on it. Fig. 8 shows the module attached to the box.

\begin{figure}[!t]
    \centering
    \includegraphics[width=0.99\columnwidth]{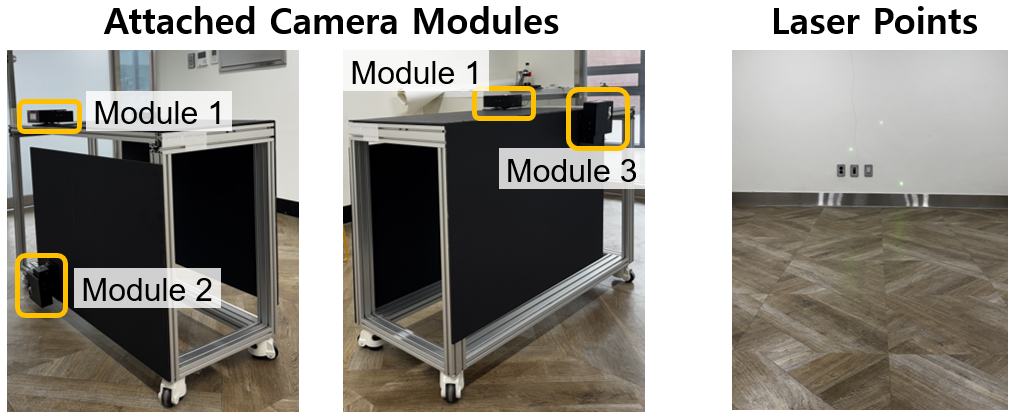}
    \caption{Camera modules attached to the frame and three laser pointers projected onto the wall}
    \label{fig:placeholder}
\end{figure}

The experiment verified the range of guidance provided by the module at 1 m intervals from the wall. Fig. 9 shows the indoor environment where the experiment was conducted. The module can receive video depending on the number of attached modules, and is currently configured to receive up to three. Fig. 10 shows an image of the screen being monitored at 1 m intervals. This confirmed stable reception up to 5 m.

\begin{figure}[!t]
    \centering
    \includegraphics[width=0.99\columnwidth]{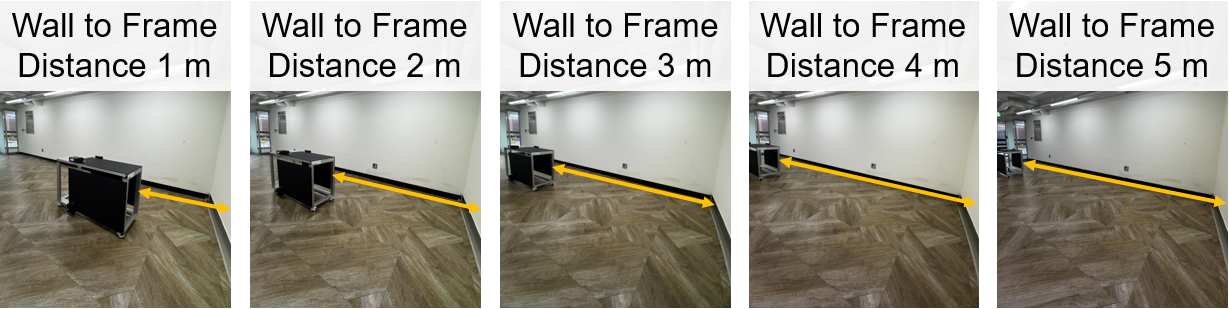}
    \caption{Experimental setup with the frame placed at 1 m intervals from the wall}
    \label{fig:placeholder}
\end{figure}

\begin{figure}[!t]
    \centering
    \includegraphics[width=0.99\columnwidth]{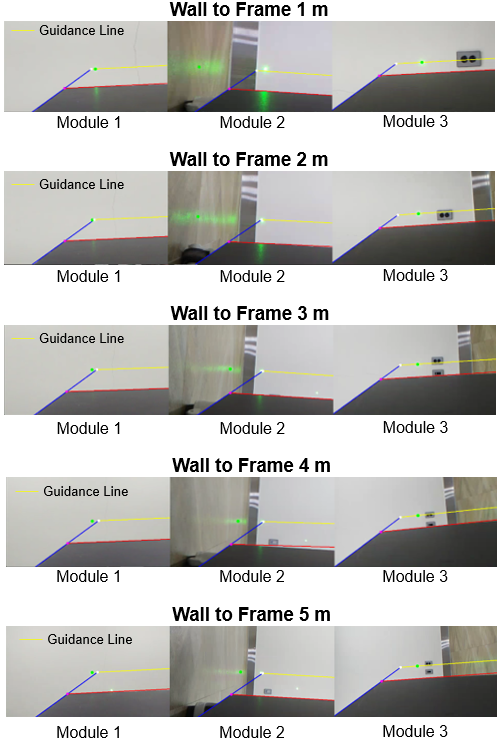}
    \caption{Experimental Results with the frame place at 1 m intervals from the wall}
    \label{fig:placeholder}
\end{figure}

Throughout the experiment, it was evident that guidance was clearly presented. It functioned normally in indoor environments, making it suitable for use. Furthermore, it was confirmed that information from three modules could be received simultaneously, providing guidance on the placement of objects. This confirmed that the modules functioned normally at heights within 5 m, demonstrating that guidance can be provided from positions higher than the height of a typical worker.

\section{Conclusion}

This study aims to prevent accidents caused by poor communication between workers during crane lowering due to the lack of information about the area directly below the load in existing crane-related machine guidance and control systems. To achieve this, a camera module was attached to the side of the load to obtain information and provide guidance on the object's placement position. When the camera module was attached to the side of the component, edge distortion in the image was used to predict the landing position on the ground, and a guidance line was drawn on the image in real time. Since the experiment did not fully represent operational environments, future work will focus on on-site validation with real crane modules.

This study focused on the camera module and functional implementation for providing guidance, so its applicability in the field is limited. Furthermore, the proposed method may have some errors due to the horizontal distance of the laser pointer from the surface of the lifted object and the deflection caused by the suction cup attachment. Therefore, in future research, we aim to apply this method to construction sites and adopt a more robust connection method to correct errors. In particular, we aim to apply this method to box-shaped components such as precast concrete modules. This is expected to contribute to improving work safety and lifting accuracy in situations where a large number of workers are directly below a relatively large object.

\end{document}